\documentclass{Interspeech}

\usepackage{CJKutf8}

\interspeechcameraready

\title{Transcript-Prompted Whisper with Dictionary-Enhanced Decoding for Japanese Speech Annotation}

\author[affiliation={1}]{Rui}{Hu}
\author[affiliation={1}]{Xiaolong}{Lin}
\author[affiliation={1}]{Jiawang}{Liu}
\author[affiliation={1}]{Shixi}{Huang}
\author[affiliation={1}]{Zhenpeng}{Zhan}

\affiliation{Baidu Inc.}{Shenzhen}{China}

\email{hurui05@baidu.com}


\usepackage{comment}

\begin{document}

\maketitle

\begin{abstract}
    
    In this paper, we propose a method for annotating phonemic and prosodic labels on a given audio-transcript pair, aimed at constructing Japanese text-to-speech (TTS) datasets. Our approach involves fine-tuning a large-scale pre-trained automatic speech recognition (ASR) model, conditioned on ground truth transcripts, to simultaneously output phrase-level graphemes and annotation labels. To further correct errors in phonemic labeling, we employ a decoding strategy that utilizes dictionary prior knowledge. The objective evaluation results demonstrate that our proposed method outperforms previous approaches relying solely on text or audio. The subjective evaluation results indicate that the naturalness of speech synthesized by the TTS model, trained with labels annotated using our method, is comparable to that of a model trained with manual annotations.
    
\end{abstract}

\section{Introduction}

The field of text-to-speech (TTS) has seen significant advancements in the past few years. Recent studies \cite{eskimez2024e2, du2024cosyvoice, liao2024fish} have demonstrated that large models trained on massive datasets can synthesize natural speech directly from grapheme sequences, eliminating the need for more fine-grained inputs such as phonemes. This is particularly notable in alphabetic languages like English. However, challenges persist for Japanese TTS due to the presence of numerous polyphones and the critical role that pitch accent plays in affecting the naturalness of speech \cite{fujimoto2019impacts, kurihara2021prosodic}.

Most TTS datasets consist solely of audio-transcript pairs, as obtaining manually annotated phonemic and prosodic labels is prohibitively expensive. As a result, many TTS systems rely on text-based frontends \cite{ying2024unified, zhang2020unified, park22b_interspeech} during data pre-processing. However, these modules can be inaccurate due to the one-to-many mapping problem between text and audio, where certain characters or words can have multiple pronunciations depending on the context. This issue is especially pronounced in the case of Japanese because: 1) there are a lot of polyphones; 2) each word has its own accent nucleus (the mora at which the pitch-downstep occurs), and adjacent words often combine to form an accent phrase and create a new accent nucleus, based on the speaker’s interpretation of the context \cite{park22b_interspeech}.

Several recent studies have explored the use of audio information during data pre-processing. Researchers in \cite{omachi2021end} proposed a sequence-to-sequence model capable of simultaneously transcribing and annotating audio with linguistic information more accurately than methods based solely on text. Building upon this, work in \cite{shirahata24_interspeech} further demonstrated the promising performance of fine-tuning a large-scale pre-trained automatic speech recognition (ASR) model to predict phonemic and prosodic labels from unlabeled speech data. However, through preliminary experiments, we found that annotation models only conditioned on audio tend to produce errors related to phonetically similar sounds. Some studies have attempted to leverage both text and audio information to obtain more precise annotations. \cite{dai22_interspeech} achieved automatic prosody labeling using text-audio data via a speech-text model while \cite{yang24_interspeech} proposed a Grapheme-to-Phoneme (G2P) module utilizing both text and aligned audio to overcome pronunciation ambiguity. However, these methods either only focus on prosody annotation or rely on complicated pipelines.

In the field of ASR, Whisper \cite{radford2023robust} is a Transformer-based \cite{vaswani2017attention} robust foundation model pre-trained on multiple tasks and large-scale data. Several recent works \cite{liao2023zero, wei24_interspeech, 10447492, jiang24b_interspeech} have demonstrated that fine-tuning Whisper with prompts can achieve impressive performance in recognizing domain-sensitive speech, spoken name entity, target speaker speech and disordered speech, respectively. 

Inspired by previous works, this paper proposes fine-tuning Whisper conditioned on ground truth transcripts to simultaneously produce phrase-level graphemes along with their phonemic and prosodic labels (hereinafter referred to as \textit{TTS labels}) from speech data. To further improve the performance of phonemic labeling, we employ a decoding strategy that leverages dictionary prior knowledge. We conduct two experiments to evaluate the effectiveness of our method. The first one is an objective evaluation. We begin by verifying that the graphemes of each phrase align faithfully with the given transcript. Then we calculate the character error rate (CER) for phonemic labeling, as well as the accuracy of accent phrase boundary prediction and the \(F_1\) scores for mora-level pitch sequence in prosodic labeling. The second one is a subjective evaluation, where human raters assess the naturalness of speech synthesized by TTS models trained on labels generated by different annotation methods. Experimental results show that the proposed method produces more accurate TTS labels and helps enhance the performance of TTS models in terms of naturalness. Audio samples are available on our demo page\footnote{https://hurui0318.github.io/japanese-annotator-demo/}.


\section{Method}

\subsection{Notation method for TTS labels}

\begin{figure*}[t]
  \centering
  \includegraphics[width=0.96\textwidth]{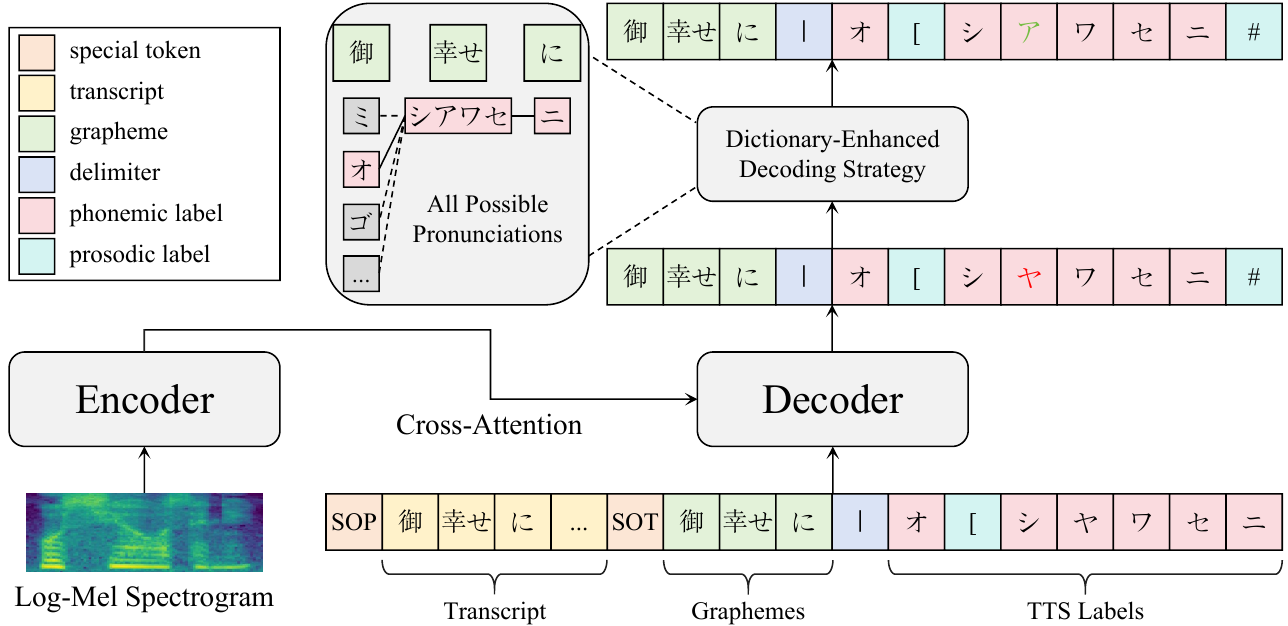}
  \caption{Overview of proposed annotation model with a decoding strategy that leverages dictionary prior knowledge.}
  \label{fig:model}
\end{figure*}

\begin{figure}[t]
  \centering
  \includegraphics[width=\linewidth]{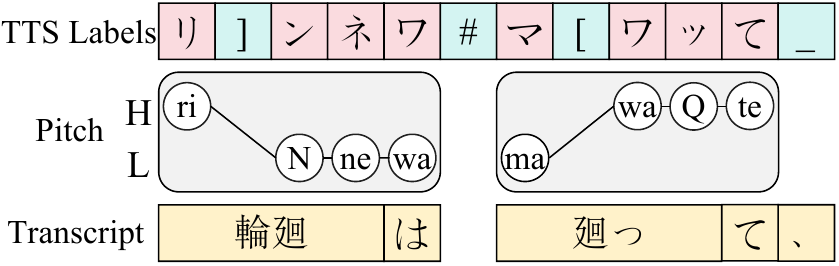}
  \caption{Example of Japanese speech annotation for \begin{CJK}{UTF8}{min}``輪廻は廻って、''\end{CJK} (The cycle of reincarnation turns, and...). The yellow squares denote the transcript segmented into accent phrases, with each accent phrase further divided into individual words. The white circles in gray area indicate the pitch status (high or low) of each mora. The red and blue squares denote phonemic and prosodic labels, respectively.}
  \label{fig:annotation}
\end{figure}

Japanese is a pitch-accent language with a writing system that combines logograms (Kanji) with phonograms (Hiragana and Katakana). As depicted in Figure~\ref{fig:annotation}, annotation for Japanese utterances typically consists of transcript and pitch status aligned at the mora level. We adopt the notation method proposed in \cite{kurihara2021prosodic} to construct TTS labels from the annotations following the rules of Tokyo dialect. The TTS labels use the following symbols to represent phonemic and prosodic information: (1) Katakana characters indicating pronunciation; (2) ``['' for pitch rising; (3) ``]'' for pitch falling; (4) ``\#'' for accent phrase boundary; (5) ``\_'' for pause.

\subsection{Annotation model}

Whisper is a robust ASR model based on the encoder-decoder Transformer architecture, pre-trained on a large amount of data. The encoder processes log-Mel spectrograms to extract high-level hidden representations, while the decoder incorporates encoder output through cross-attention mechanism and generates transcripts in an auto-regressive manner. In the decoder input, we can use a \texttt{<startofprev>} token to represent the beginning of previous context and a \texttt{<startoftranscript>} token to indicate the beginning of transcription.

Researchers in \cite{shirahata24_interspeech} first proposed fine-tuning pre-trained Whisper model to produce TTS labels conditioned on audio. However, in our preliminary experiments, we observed that the model tends to lose some semantic knowledge acquired during pre-training. Consequently, it is prone to making mistakes such as failing to distinguish between phonetically similar sounds, omitting or adding certain pronunciations. Some examples are shown in Table~\ref{tab:badcases}. We speculate that this is likely because the model relies solely on speech as input, with training targets limited to TTS labels. To mitigate this, inspired by \cite{omachi2021end}, we also attempted to have the model simultaneously output phrase-level graphemes and TTS labels, expecting that the model would understand the speech content before annotation. We found that the model has shown some improvements but errors still persist. This is reasonable since the base model, Whisper, is not a perfect transcriber. Therefore, during inference, the preceding graphemes may be inaccurate and could provide insufficient, even misleading guidance for the subsequent decoding of TTS labels.

\begin{table}[t]
    \caption{Examples of errors produced by the annotation model. ``Content'' denotes the ground truth transcript of speech content. ``Reference'' and ``Prediction'' refer to the annotations from the human annotator and the annotation model, respectively.}
    \label{tab:badcases}
    \centering
    \begin{tabular}{ll}
        \toprule
        \textbf{Source} & \textbf{Example} \\
        \midrule
        Content & \begin{CJK}{UTF8}{min}成功してもしなく\textcolor{blue}{て}も\end{CJK} \\
        Reference & \begin{CJK}{UTF8}{min}セ[ーコーシテ]モ\#シ[ナ]ク\textcolor{blue}{テ}モ\#\end{CJK} \\
        Prediction & \begin{CJK}{UTF8}{min}セ[ーコーシテ]モ\#シ[ナ]ク\textcolor{red}{タ}モ\#\end{CJK} \\
        \midrule
        Content & \begin{CJK}{UTF8}{min}\textcolor{blue}{漁}業と商業\end{CJK} \\
        Reference & \begin{CJK}{UTF8}{min}\textcolor{blue}{ギョ}]ギョート\#ショ]ーギョーデ\#\end{CJK} \\
        Prediction & \begin{CJK}{UTF8}{min}\textcolor{red}{ギョ}]\textcolor{red}{ー}ギョート\#ショ]ーギョーデ\#\end{CJK} \\
        \midrule
        Content & \begin{CJK}{UTF8}{min}新しい\textcolor{blue}{家}\end{CJK} \\
        Reference & \begin{CJK}{UTF8}{min}ア[タラシ]ー\#\textcolor{blue}{イ}[\textcolor{blue}{エ}]オ\#\end{CJK} \\
        Prediction & \begin{CJK}{UTF8}{min}ア[タラシ]ー\#\textcolor{red}{エ}]オ\#\end{CJK} \\
        \bottomrule
    \end{tabular}
\end{table}

Our proposed annotation model, however, takes both audio and its transcript as input and is required to output phrase-level graphemes along with their TTS labels. As illustrated in Figure~\ref{fig:model}, the transcript is tokenized and placed between \texttt{<startofprev>} and \texttt{<startoftranscript>} as a prompt to the decoder. Symbol ``\textbar'' is the delimiter between graphemes and TTS labels of each phrase. In this approach, the decoder can attend to both the acoustic information from input speech and the semantic information of the transcript when generating labels.

\subsection{Decoding strategy}\label{sec:decoding}

Upon close examination, we still found the aforementioned errors present in the phonemic labeling of Kanji characters. Since the model is requested to sequentially output the graphemes and TTS labels of each accent phrase during decoding, we can utilize the preceding graphemes and a dictionary to correct the following phonemic labels.

As shown in Figure~\ref{fig:model}, when the annotation mode decodes an accent phrase boundary symbol ``\#'', we first get the graphemes and TTS labels for the phrase based on the delimiter ``\textbar''. The predicted TTS labels are converted into pronunciation in Katakana and accent type (the position of mora where accent nucleus appears) following the rules of Tokyo dialect. Then we use MeCab \cite{kudo2004applying} with UniDic \cite{den2008proper} to segment the graphemes into words and look up each word's candidate pronunciations to get all possible pronunciations for the phrase. Next, we select the one with the shortest edit distance to the pronunciation predicted by the annotation model as the corrected pronunciation. After this, we need to restore it to the form of TTS labels with some modifications to the accent type. As shown in Equation~\ref{eq:restoration}, the accent type remains unchanged for phrases whose mora size stays the same after pronunciation correction or when the original accent type is \(0\) (flat), \(1\) (head high) or \(n\) (tail high), where \(n\) is the mora size. For the rest (middle high), the modified accent type \(A_{\text{mod}}\) is calculated by adding the difference between the mora size of the corrected phrase \(M_{\text{mod}}\) and that of the original phrase \(M_{\text{orig}}\) to the original accent type \(A_{\text{orig}}\). Finally, we replace the original TTS labels with the corrected ones and continue with the decoding of subsequent phrases.

\begin{align}
    A_{\text{mod}} = 
    \begin{cases}
        A_{\text{orig}}, & A_{\text{orig}} \in \{0, 1, n\}, \\
        A_{\text{orig}} + (M_{\text{mod}} - M_{\text{orig}}), & A_{\text{orig}} \in \{2, \dots, n-1\}.
    \end{cases}
    \label{eq:restoration}
\end{align}

Note that the restoration of TTS labels is not perfect but we show that the impact is minimal in the subsequent experiments.

\section{Experimental setup}

\subsection{Evaluation metrics}

In the experiments, we aim to evaluate our proposed annotation method in three aspects:

\begin{itemize}
    \item The alignment of predicted phrase-level graphemes with the given transcript.
    \item The performance of phonemic and prosodic labeling.
    \item The effectiveness of its application in TTS systems.
\end{itemize}

We first segment the predicted utterance-level TTS labels into accent phrases based on the accent phrase boundary symbol ``\#''. For each accent phrase, we divide it into graphemes and TTS labels using the delimiter ``\textbar''. Following the rules of Tokyo dialect, the phrase-level TTS labels are converted into phonemic labels in Katakana and mora-level high-low pitch sequences as prosodic labels. The phrase-level graphemes, phonemic labels and prosodic labels are then concatenated back to the utterance level, respectively. CERs are calculated for both the grapheme sequences and phonemic labels. 

Utterances where all models correctly predict the phonemic labels are used for evaluating prosodic labeling. This is because comparing ground truth prosodic labels with predicted ones becomes infeasible when the predicted phonemic labels are incorrect \cite{shirahata24_interspeech}. To evaluate the accuracy of accent phrase boundary prediction, we calculate a metric defined as the number of correctly predicted accent phrases divided by the total number of accent phrases in the manual annotations. For pitch labeling, we compute \(F_1\) scores on the mora-level high-low pitch sequences.

 Finally, we directly use the utterance-level TTS labels produced by the annotation model to train TTS models and conduct subjective listening test based on the mean opinion score (MOS). Ten raters proficient in Japanese are asked to assess 50 randomly selected speech samples for pronunciation accuracy and prosodic naturalness.

\subsection{Data}

We use three datasets for model training and evaluation:
\begin{description}
    \item[JSUT: ] a public Japanese speech corpus with a single female voice \cite{sonobe2017jsut}. The {\it basic5000} subset and their manually annotated TTS labels\footnote{https://github.com/sarulab-speech/jsut-label} are used. The TTS labels are at the accent phrase level but there is no alignment information with the text. Therefore, we feed them into GPT-4 \cite{achiam2023gpt} to obtain phrase-level aligned annotations. The data are used for training annotation models.
    \item[PRIV: ] a private Japanese speech corpus recorded by seven male and four female professional native Japanese speakers. It comprises 104,183 text samples and 121.82 hours of speech, with annotations provided at the mora level. Data from six male and four female speakers are used to train the annotation models, while the remaining data from one male speaker, consisting of 8,888 samples, are reserved for testing. The held-out data are further randomly split into 7,110, 889 and 889 samples for training, validation and testing in TTS experiments, respectively.
    \item[JVS: ] a public Japanese speech corpus containing audio-transcript pairs of 100 professional speakers \cite{takamichi2019jvs}. We only use the {\it parallel100} and {\it nonpara30} subsets, with a total of 13,000 samples. These samples are randomly divided into 10,400, 1300 and 1300 samples for training, validation and testing in TTS experiments, respectively.
    
\end{description}

\subsection{Model details}


The proposed annotation model is fine-tuned from {\it whisper-large-v3-turbo}\footnote{https://huggingface.co/openai/whisper-large-v3-turbo} for its robustness and inference speed, using the transformers library \cite{wolf-etal-2020-transformers}. The model is fine-tuned for 100k steps with a batch size of 32 on a single A100 GPU. The learning rate is initialized at 0.00001 with 500 warm-up steps. During training, the labels for the prompt tokens are set to \(-100\) to ignore loss calculation. Notably, in our preliminary experiments, we found that directly using symbols, such as ``\#'' could lead to severe hallucinations during inference. We conjecture that this is because these symbols have already been assigned different meanings during pre-training, which are difficult to alter through fine-tuning. Therefore, we use Unicode characters U+2191, U+2193, U+2460, U+2462 and U+2223, which are not present in the Whisper tokenizer's vocabulary, to represent ``['', ``]'', ``\#'', ``\_'' and ``\textbar'', respectively. The following summarizes the annotation methods used in our experiments:
\begin{description}
    \item[NLP: ] Open JTalk\footnote{https://open-jtalk.sp.nitech.ac.jp/}, a text-based annotation method. We use its Python wrapper\footnote{https://github.com/r9y9/pyopenjtalk}.
    \item[Annt-v1: ] Whisper model that is conditioned solely on audio and is fine-tuned to output TTS labels. We train two models on JSUT and PRIV, referred to as \textbf{Annt-v1-pub} and \textbf{Annt-v1-priv}.
    \item[Annt-v2: ] Whisper model that is conditioned solely on audio and is fine-tuned to outupt phrase-level graphemes and TTS labels. We train two models on JSUT and PRIV, referred to as \textbf{Annt-v2-pub} and \textbf{Annt-v2-priv}. 
    \item[Annt-v3: ] Proposed method where Whisper model is conditioned on both audio and its corresponding transcript, fine-tuned to output phrase-level graphemes and TTS labels. We train two models on JSUT and PRIV, referred to as \textbf{Annt-v3-pub} and \textbf{Annt-v3-priv}.
    \item[Annt-v4: ] \textbf{Annt-v3} incorporating proposed decoding strategy. We apply it to both \textbf{Annt-v3-pub} and \textbf{Annt-v3-prv}, referred to as \textbf{Annt-v4-pub} and \textbf{Annt-v4-priv}.
\end{description}

The TTS model is built upon VITS \cite{kim2021conditional}, an end-to-end architecture that can synthesize high-quality speech. We use the official implementation\footnote{https://github.com/jaywalnut310/vits} and train all models for 500k steps on a single A100 GPU, adhering to the original configuration, using TTS labels generated by the text-based method and annotation models trained on PRIV. Additionally, we incorporate two more sources to ensure a fairer comparison:
\begin{description}
    \item[Reference: ] Recorded speech samples.
    \item[Oracle: ] TTS model trained on manually annotated labels. Since JVS lacks manual annotations, we only train a model on PRIV.
\end{description}

For each TTS model, we input the labels produced by its corresponding annotation method to generate speech samples for evaluation, ensuring consistency between training and inference.

\section{Experimental results}

We calculate the CER of grapheme sequences predicted by \textbf{Annt-v3-priv} on the testing data of PRIV and the result is 0.16\%. Interestingly, most of the errors occur when the model changes graphemes to another form identical in pronunciation, such as \begin{CJK}{UTF8}{min}``ブィ''\end{CJK} and \begin{CJK}{UTF8}{min}``ビ''\end{CJK}, or when it normalizes numbers, such as converting \begin{CJK}{UTF8}{min}``1週''\end{CJK} to \begin{CJK}{UTF8}{min}``一週''\end{CJK}. One possible reason is the inconsistency of annotation style across human annotators. Therefore, we can assume that the predicted graphemes of each accent phrase align well with the given transcript, ensuring the feasibility of our proposed decoding strategy. 

\begin{table}[t]
    \caption{Objective evaluation results on each metric. CER is for phonemic labeling while accuracy and $F_1$ are for prosodic labeling}
    \label{tab:objective_evaluation}
    \centering
    \begin{tabular}{lccc}
        \toprule
        \textbf{Model} & \textbf{CER ($\downarrow$)} & \textbf{Acc. ($\uparrow$)} & $\mathbf{F_1}$ \textbf{($\uparrow$)} \\
        \midrule
        NLP & 3.56\% & 69.33\% & 83.69\% \\
        \midrule
        Annt-v1-pub & 2.41\% & 81.49\% & 92.03\% \\
        Annt-v2-pub & 1.92\% & 81.79\% & 92.15\% \\
        Annt-v3-pub (ours) & 1.59\% & \textbf{81.95\%} & \textbf{92.27\%} \\
        Annt-v4-pub (ours) & \textbf{1.31\%} & \textbf{81.95\%} & 92.25\% \\
        \midrule
        Annt-v1-priv & 1.13\% & 86.04\% & 94.35\% \\
        Annt-v2-priv & 0.80\% & 86.40\% & 94.55\% \\
        Annt-v3-priv (ours) & 0.63\% & \textbf{87.32\%} & \textbf{94.59\%} \\
        Annt-v4-priv (ours) & \textbf{0.57\%} & \textbf{87.32\%} & 94.58\% \\
        \bottomrule
    \end{tabular}
\end{table}

Table~\ref{tab:objective_evaluation} shows the objective evaluation results on the PRIV testing data. All annotation models utilizing audio information significantly outperform the text-based method (i.e., \textbf{NLP}). Among the models conditioned solely on audio, those that produce both phrase-level graphemes and TTS labels (i.e., \textbf{Annt-v2-pub} and \textbf{Annt-v2-priv}) perform better than those that only generate TTS labels (i.e., \textbf{Annt-v1-pub} and \textbf{Annt-v1-priv}). Our proposed models (i.e., \textbf{Annt-v3-pub} and \textbf{Annt-v3-priv}), relying on audio and the corresponding transcript, are more accurate in terms of both phonemic and prosodic labeling. The models applying proposed decoding strategy (i.e., \textbf{Annt-v4-pub} and \textbf{Annt-v4-priv}) achieve the best performance of phonemic labeling with a slight degradation in pitch labeling, which is expected as mentioned in Section \ref{sec:decoding}. The accuracy of accent phrase boundary prediction remains unchanged since the decoding strategy is performed within each accent phrase.

\begin{table}[t]
    \caption{MOS test results for TTS models trained using TTS labels from different annotation models with 95\% confidence intervals.}
    \label{tab:subjective_evaluation}
    \centering
    \begin{tabular}{lcc}
        \toprule
        \textbf{Model}	& \textbf{JVS} &  \textbf{PRIV} \\
        \midrule
        Reference & 4.43 \(\pm\) 0.08 & 4.68 \(\pm\) 0.07 \\
        Oracle & - & 4.27 \(\pm\) 0.08 \\
        NLP & 3.63 \(\pm\) 0.09 & 3.85 \(\pm\) 0.10 \\
        Annt-v1-priv & 3.87 \(\pm\) 0.08 & 4.06 \(\pm\) 0.09 \\
        Annt-v2-priv & 3.92 \(\pm\) 0.09 & 4.11 \(\pm\) 0.09 \\
        Annt-v3-priv (ours) & 4.12 \(\pm\) 0.09 & 4.23 \(\pm\) 0.09 \\
        Annt-v4-priv (ours) & \textbf{4.16 \(\pm\) 0.08} & \textbf{4.29 \(\pm\) 0.08} \\
        \bottomrule
    \end{tabular}
\end{table}

Table~\ref{tab:subjective_evaluation} presents the results of subjective evaluation. Consistent with the objective evaluation results, all methods leveraging audio information significantly outperform the text-based method. Furthermore, our proposed method not only outperforms the baseline methods across all datasets but also achieves scores slightly higher than manual annotation (i.e., \textbf{Oracle}) on PRIV. This is probably due to the inconsistent annotation style across human annotators, which is also reported in \cite{shirahata24_interspeech, dai22_interspeech}.
The results demonstrate that the proposed method is capable of generating high-quality annotations from audio-transcript pairs, thereby enhancing the performance of Japanese TTS models.

\section{Conclusion}

In this paper, we propose an annotation model fine-tuned from a pre-trained ASR model to produce graphemes along with their phonemic and prosodic labels for given audio-transcript pairs. To address the phonemic labeling issue associated with Kanji characters, we adopt a decoding strategy that incorporates dictionary prior knowledge. Experimental results demonstrate that the proposed annotation method is effective in building high-quality Japanese TTS models. Since the model can generate phrase-level graphemes and TTS labels, it holds potential for constructing polyphone disambiguation datasets and training text-based accent sandhi estimation models in future work.

\newpage
\bibliographystyle{IEEEtran}
\bibliography{mybib}

\end{document}